\pdfoutput=1
\documentclass[11pt]{article}

\usepackage[final]{acl}

\usepackage{times}
\usepackage{latexsym}

\usepackage[utf8]{inputenc}
\usepackage[T1]{fontenc}
\usepackage{CJKutf8}  
\usepackage{microtype}
\usepackage{inconsolata}

\usepackage{graphicx}
\usepackage{amssymb}
\usepackage{amsmath}

\usepackage{multirow}
\usepackage{booktabs}

\usepackage{subcaption}
\usepackage{csvsimple-l3}
\usepackage{pgfplotstable}
\usepackage{longtable}
\usepackage{geometry}
\usepackage{graphicx}
\usepackage{adjustbox}
\usepackage{float}

\title{Massive Supervised Fine-tuning Experiments Reveal How Data, Layer, and Training Factors Shape LLM Alignment Quality}

\author{
 \textbf{Yuto Harada\textsuperscript{1,2*}}\thanks{Equal Contribution.},
 \textbf{Yusuke Yamauchi\textsuperscript{1,2*}}\footnotemark[1],
 \textbf{Yusuke Oda\textsuperscript{1,3}},
 \textbf{Yohei Oseki\textsuperscript{1,2}},
 \\
 \textbf{Yusuke Miyao\textsuperscript{1,2}}\thanks{Corresponding authors: Yusuke Miyao and Yu Takagi},
 \textbf{Yu Takagi\textsuperscript{4}}\footnotemark[2]
\\
 \textsuperscript{1}NII LLMC,
 \textsuperscript{2}The University of Tokyo,
 \textsuperscript{3}NAIST,
 \textsuperscript{4}Nagoya Institute of Technology,
\\
\texttt{\{harada-yuto, yamauchi\_y\}@nii.ac.jp} \\
\texttt{yusuke@is.s.u-tokyo.ac.jp} \quad
\texttt{takagi.yu@nitech.ac.jp}
}

\begin{document}    
\maketitle

\begin{abstract}
\label{sec:abstract}
Supervised fine-tuning (SFT) is a critical step in aligning large language models (LLMs) with human instructions and values, yet many aspects of SFT remain poorly understood. 
We trained a wide range of base models on a variety of datasets including code generation, mathematical reasoning, and general-domain tasks, resulting in 1,000+ SFT models under controlled conditions. We then identified the dataset properties that matter most and examined the layer-wise modifications introduced by SFT.
Our findings reveal that some training–task synergies persist across all models while others vary substantially, emphasizing the importance of model-specific strategies. Moreover, we demonstrate that perplexity consistently predicts SFT effectiveness, often surpassing superficial similarity between the training data and the benchmark, and that mid-layer weight changes correlate most strongly with performance gains. We release these 1,000+ SFT models and benchmark results to accelerate further research. All resources are available at \url{https://github.com/llm-jp/massive-sft}.
\end{abstract}    
\section{Introduction}
\label{sec:introduction}

Recent advances in large language models (LLMs) have greatly improved natural language understanding and generation. However, pretrained LLMs often fail to align with human intentions or specific tasks~\citep{ouyang2022training}, motivating alignment methods. 
Supervised fine-tuning (SFT) trains models to follow human instructions and remains a widely used and effective approach for improving downstream performance~\citep{weifinetuned, guan2024deliberative}.

Although recent works have explored how model size and training-data characteristics influence downstream tasks in the context of SFT~\citep{jin2024demystifying, dong-etal-2024-abilities}, large-scale research specifically examining which aspects of SFT datasets benefit different base models remains limited.
While some studies compare or analyze publicly available models~\citep{oyama2025mapping1000languagemodels}, these are not controlled experiments and often introduce biases, such as favoring certain model families. Consequently, it remains unclear how SFT of various models on different datasets affects benchmark performance, how relationships among datasets and benchmarks vary across models, and which internal weights are most responsible for these effects. Furthermore, there are several SFT training approaches including Low-Rank Adaptation (LoRA)~\citep{hu2022lora}, and there is ongoing debate about the optimal amount of data required~\citep{zhou2024lima, chen2023alpagasus}; however, there has yet to be a comprehensive, quantitative comparison. Hence, a comprehensive examination of these issues on SFT is urgently needed.

\begin{figure*}[t]
\centering
% \framebox[\textwidth]{\rule{0pt}{5cm}\centering Placeholder for the image}
\includegraphics[width=1.0\linewidth]{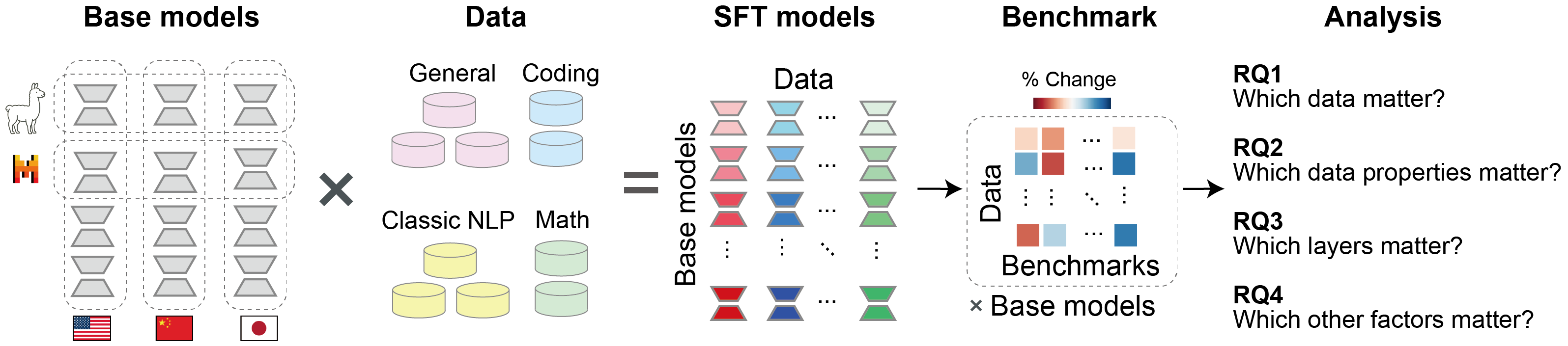}
\caption{Overview of this study. We conduct SFT on numerous combinations of base models and training data.  These models are evaluated on a variety of benchmark tasks to comprehensively examine the relationships among the base models, training data, and benchmark tasks.}
\label{fig:overview}
\end{figure*}

In this study, we trained twelve base models on multi-domain datasets, produced a large suite of SFT models, and evaluated them across diverse tasks (Figure~\ref{fig:overview}). Specifically, we address the following Research Questions (RQs):

\begin{enumerate}
    \item How do models, training data, and benchmarks interact for downstream performance? Do any training datasets yield consistent gains across models, or are improvements model-specific? Are relationships between datasets and benchmarks stable across models?
    % \item How do models, training data, and benchmarks interact with one another? Do certain training datasets consistently enhance benchmark performance across a variety of models, or does each model exhibit its own distinct preferences? Likewise, do relationships among different datasets and benchmarks remain the same across models?
    \item Which properties of the training data used for SFT affect downstream performance?
    \item Which layers in the model are most critical for SFT? Are there universal patterns across different models?
    \item How do factors debated in SFT, including training method, sample size, and cross-lingual transfer, relate to performance?
\end{enumerate}
In summary, our contributions are as follows:
% The main contributions of this work can be summarized as follows:

\paragraph*{Large-Scale, Integrated Evaluation}  
By systematically performing SFT on multiple base models and various training datasets, we uncover the complexity of relationships among models, data, and downstream tasks. While the relationships between training data and evaluation tasks follow broadly similar patterns across models, they also exhibit model-specific characteristics.

\paragraph*{Revealing a Simple ``Perplexity Is Key'' Law}  
We find that training data with lower perplexity for the base model consistently leads to greater improvements in downstream performance. In contrast, factors once considered crucial, such as content similarity between training and evaluation data or tokenizer compatibility, do not exhibit as strong an effect as perplexity.

\paragraph*{Strong Correlation Between Mid-Layer Weight Changes and Performance}  
We observe that changes in mid-layer weights correlate more strongly with downstream performance gains than changes in either the top or bottom layers. Indeed, intrinsic dimensionality analysis of embeddings revealed that the embedding space begins to diverge substantially from the base model at mid-layer positions, suggesting these layers actively expand the model’s representational subspace during SFT. This pattern appears consistent across multiple models, offering critical insights for efficient fine-tuning and model monitoring.

\paragraph*{Embedding the SFT Landscape}
Projecting the log-likelihood vectors of fine-tuned models into a common latent space lets us compare diverse training dynamics in one coordinate system. The resulting map shows that the global layout is determined by model family rather than training corpus, that checkpoints from successive epochs converge toward a shared instruction-following region, that enlarging the instruction set from 1k to 20k nudges models only slightly outward from this center, and that LoRA trajectories almost perfectly overlap those of full-parameter tuning.

\paragraph*{Resource Release for Future Research}  
All fine-tuned models produced in this study are publicly released. We expect this comprehensive set of models to accelerate deeper investigations of SFT and to foster rapid progress in the field.

\section{Related Work}
\label{sec:related-work}
The role of training data characteristics in SFT has been highlighted in many prior studies. For instance, mixing code-generation data has been suggested to enhance a model’s reasoning and logical abilities \citep{dong-etal-2024-abilities}. Similarly, incorporating instruction data that includes procedural knowledge could improve mathematical reasoning \citep{ruis2024procedural}. Furthermore, considering task relevance when selecting datasets can lead to more robust performance \citep{huang2024o1,zhang2024unveiling}. 

While early work focused on how to fine-tune, comparing full-parameter updates against LoRA \citep{ivison2023camels,zhuo2024astraios, dettmers2024qlora,zhao2024galore,biderman2024lora}, or debating sample size \citep{zhou2024lima, zhao2024long, chen2023alpagasus}. More recent studies have shifted attention to the statistics of the training data itself. For example, \citet{jin2024demystifying} and \citet{wu2025clearmindsthinkalike} independently show that lower perplexity and moderate sequence length are stronger predictors of SFT success than sheer volume.

Overall, most studies focus on particular models or tasks, and there remains a lack of comprehensive, large-scale evaluations across multiple models. This study aims to offer a broader perspective by controlling for model, data, and fine-tuning methods on a larger scale, thus providing more integrated insights into SFT behavior.

\section{Methods}
\label{sec:methods}
This section describes the base models, SFT procedures, and evaluation benchmarks.

\subsection{Base Models}
We employed a total of 12 models with approximately 7B parameters each across English, Chinese, and Japanese for SFT experiments. Specifically, we selected \textbf{English models}: OLMo-7B \citep{Groeneveld2023OLMo}, Llama3-8B \citep{dubey2024llama}, Mistral-7B \citep{jiang2023mistral7b}, and Gemma2-9B \citep{gemmateam2024gemma2improvingopen}; \textbf{Chinese models}: Qwen2.5-7B \citep{qwen2}, Chinese-Llama3-8B \citep{chinese-llama-alpaca},
% chinese-llama-3: 論文内に3は出ていないが https://github.com/ymcui/Chinese-LLaMA-Alpaca-3/blob/main/README_EN.md?utm_source=chatgpt.com ではこの論文を引用するように言われている
Chinese-Mistral-7B \citep{hsu2024breeze7btechnicalreport}, and Yi1.5-9B \citep{ai2025yiopenfoundationmodels};
 % Yi-1.5: 1.5には論文がないが、https://github.com/01-ai/Yi-1.5?tab=readme-ov-fileでPaperをクリックすると1のpaperに飛ぶ
and \textbf{Japanese models}: LLMjp-3-7B \citep{llmjp2024llmjpcrossorganizationalprojectresearch},
 % llm-jp-3: 3シリーズはモデルカードのみ
Llama3-Swallow-8B \citep{fujii2024continualpretrainingcrosslingualllm}, Swallow-Mistral-7B \citep{fujii2024continualpretrainingcrosslingualllm}, and Sarashina2-7B\footnote{https://huggingface.co/sbintuitions/sarashina2-7b}.
 % Sarashina: https://www.sbintuitions.co.jp/blog/entry/2024/06/26/115641しかレポートが見当たらない 
By comparing these diverse models, we investigate not only cross-lingual differences but also behaviors during continual pretraining within model families such as the Llama family (Llama3, Chinese-Llama3, Llama3-Swallow) and the Mistral family (Mistral, Chinese-Mistral, Swallow-Mistral).
To facilitate fair comparison at the peak effectiveness of instruction-tuning, all base models used in this experiment had not undergone any subsequent post-training.
More information on each model can be found in Appendix~\ref{sec: description of base models}.

\subsection{Training Datasets}
\label{sec: Training Datasets}
We utilized 10 distinct datasets categorized into 4 major groups. Although our base models cover English, Chinese, and Japanese, all training datasets used for SFT are exclusively in English. Specifically, we selected \textbf{General Tasks}: Alpaca \citep{alpaca}, LIMA \citep{zhou2024lima}, and UltraChat \citep{ding2023enhancing}; \textbf{Coding Tasks}: CodeAlpaca \citep{codealpaca} and Magicoder \citep{wei2024magicoder}; \textbf{Math Tasks}: OpenMathInstruct \citep{toshniwal2024openmathinstruct} and MathInstruct \citep{yue2023mammoth}; and \textbf{Classic NLP Tasks}: FLAN \citep{weifinetuned}.
The FLAN dataset further consists of 3 subcategories. 
FLAN Knowledge includes BoolQ \citep{clark2019boolq}, NaturalQuestions \citep{47761}, and TriviaQA \citep{joshi-etal-2017-triviaqa}. 
FLAN Reasoning includes ARC-Easy \& Challenge \citep{allenai:arc}, HellaSwag \citep{zellers2019hellaswag}, WinoGrande \citep{sakaguchi2019winogrande}, and PIQA \citep{Bisk2020}. 
FLAN Comprehension includes QuAC \citep{choi-etal-2018-quac} and SQuAD v2 \citep{rajpurkar-etal-2018-know}.
The categorization of FLAN follows the criteria defined in \citet{dubey2024llama, 2023opencompass}.

To uniformly compare a wide variety of base models, all datasets were preprocessed under consistent conditions. Initially, samples exceeding the maximum sequence length supported by all models' tokenizers were removed, as overly long samples cannot be adequately learned. Subsequently, either 1k or 20k samples were randomly extracted from each dataset.
Further details on the training datasets are provided in Appendix~\ref{sec: description of training datasets}.

\subsection{Training Settings}
We trained a total of 1,070 models by varying several conditions. First, all 12 models underwent both full-parameter and LoRA training with a sample size of 1k for each individual dataset. Additionally, we conducted training using a combined dataset (All Dataset) to assess the effect of mixing all data.

For further validation, we conducted additional experiments using 3 primary models (OLMo, Qwen, and LLM-jp), focusing on the impact of dataset size by comparing training results using 1k and 20k samples. In this specific experiment, the learning rate schedule was switched from cosine (used in regular training) to constant to isolate the effect of dataset size.

Through preliminary experiments, we determined optimal hyperparameters for both full-parameter fine-tuning and LoRA, ensuring that the supervised fine-tuning process was conducted under stable and well-tuned conditions.
Details of the preliminary experiments are provided in Appendix~\ref{sec: preliminary experiments}, while training configurations, computational costs, and a few exceptional cases where training did not complete successfully are described in Appendix~\ref{sec: description of training settings}.

\begin{figure*}[t]
    \centering
    % \fbox{\rule{0.95\linewidth}{8cm}}%
    \includegraphics[width=1.0\linewidth]{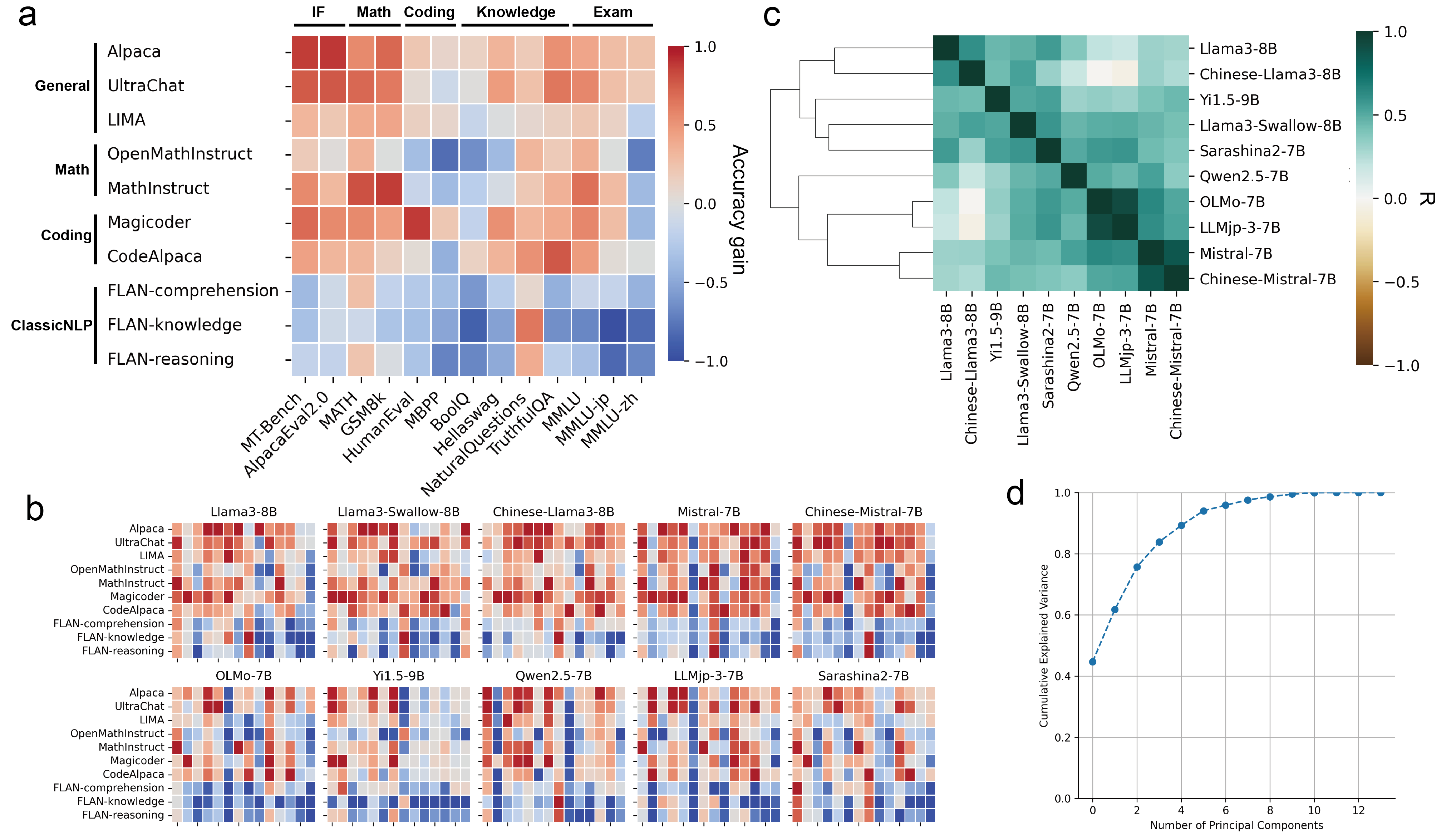}
    \caption{
    \textbf{a} Average of the performance change for diverse benchmarks from the each baseline model after SFT on each training dataset. Each column is min-max scaled to the [$-1,1$] range.
    \textbf{b} The performance changes visualized for each model individually.
    \textbf{c} Pairwise correlation matrix of performance changes across all SFT models, with the corresponding hierarchical-clustering dendrogram superimposed.
    \textbf{d} The cumulative explained variance ratio obtained by applying PCA to all concatenated results from \textbf{b}.
    }
    \label{fig:train-task-accuracy}
\end{figure*}

\subsection{Evaluation}
We evaluated all models on downstream tasks using OpenCompass\footnote{We used the GitHub repository from OpenCompass: \url{https://github.com/open-compass/opencompass}} \citep{2023opencompass}, a large-scale evaluation tool.
We evaluated model performance across 12 benchmark datasets spanning 5 categories: covering \textbf{Math} (MATH \citep{MATH}, GSM8K \citep{GSM8k}), \textbf{Coding} (HumanEval \citep{HumanEval}, MBPP \citep{MBPP}), \textbf{Knowledge} (BoolQ \citep{clark2019boolq}, NaturalQuestions \citep{NQ}, TruthfulQA \citep{TruthfulQA}), \textbf{Examination} (MMLU \citep{MMLU1, MMLU2}, MMLU-zh \citep{CMMLU}, MMLU-jp) and \textbf{Instruction-following} (MT-Bench \citep{MTBench}, AlpacaEval v2.0 \citep{AlpacaEvalv2}). A detailed description is provided in the Appendix \ref{sec: description of the dataset}. As all models were trained in a zero-shot instruction-response format, we focus primarily on zero-shot inference results in our evaluation. Gemma2-9B and Swallow-Mistral-7B were excluded due to inconsistent evaluation conditions, and we report results mainly for the remaining 10 models.

\section{Results}
\label{sec:results}

\subsection{RQ1. Relationship Among Models, Training Data, and Downstream Tasks}

\begin{figure*}[t]
    \centering
    % \fbox{\rule{0.95\linewidth}{5cm}}%
    \includegraphics[width=1.0\linewidth]{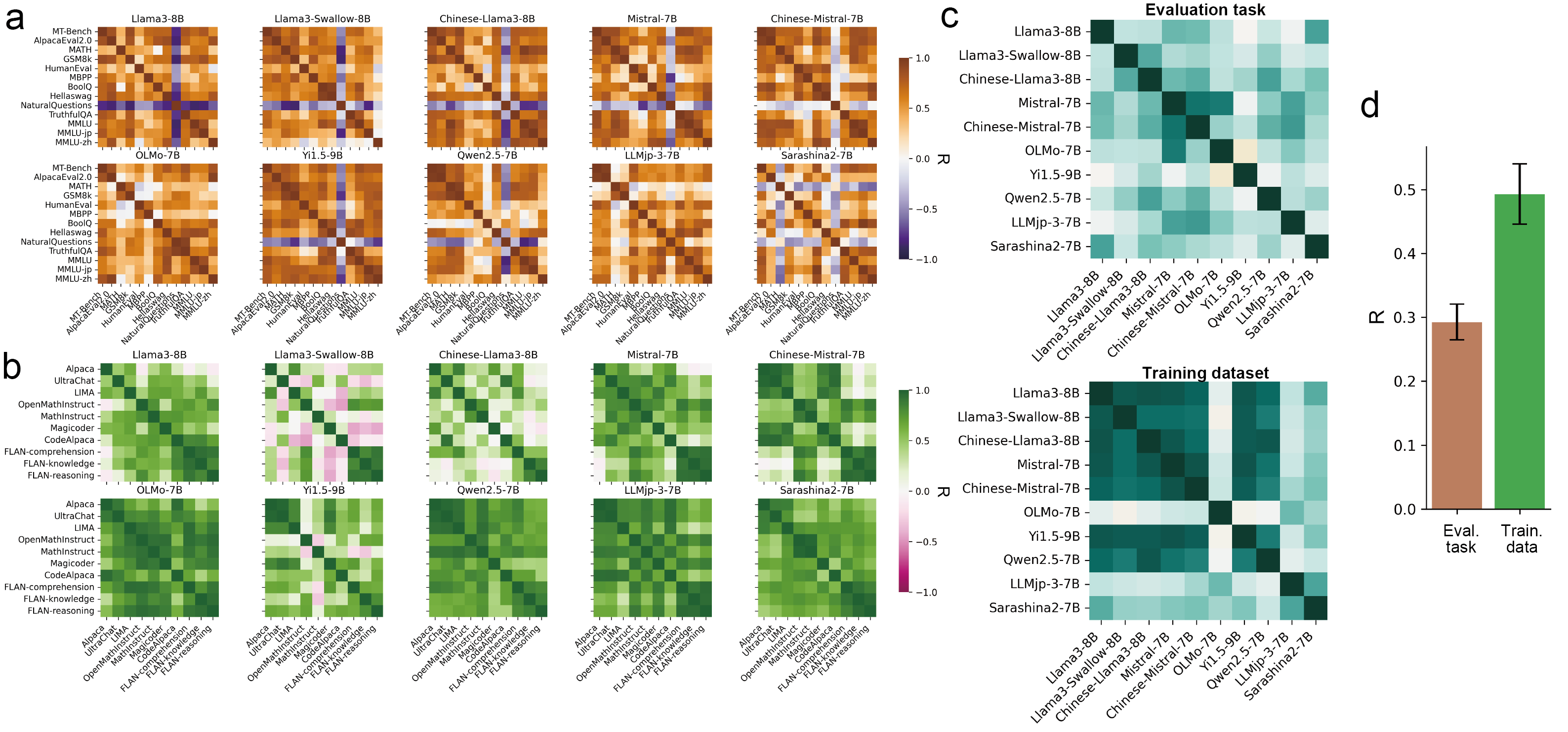}
    \caption{
    \textbf{a} Pairwise correlations between evaluation tasks in terms of performance improvements across training datasets.
    \textbf{b} Similar to \textbf{a}, but focusing on relationship between correlations between training datasets.  
    \textbf{c} Model-to-model similarity for \textbf{a} (top)  and \textbf{b} (bottom), respectively.  
    \textbf{d} Comparison of the lower-triangle elements of the two similarity matrices in \textbf{c}.
    }
    \label{fig:train-task-correlation}
\end{figure*}

First, we examine how various base models interact with different training datasets and how these relationships shape downstream performance. We aim to determine whether certain datasets provide uniform benefits across models or if each model exhibits unique sensitivities. To this end, we analyze evaluation results obtained by fine-tuning each of the ten base language models with each of the ten SFT training datasets, every dataset containing 1k examples.

Figure~\ref{fig:train-task-accuracy}a visualizes the relationship between training datasets and downstream tasks when aggregating results across all models. Some datasets show clear improvements for multiple tasks, while others offer minimal, or even negative gains. For instance, Alpaca and UltraChat generally deliver consistent performance boosts, whereas FLAN is detrimental to most tasks (except Natural Questions, which aligns with its domain). In addition, MathInstruct and OpenMathInstruct particularly boost MATH and GSM8K, whereas Magicoder benefits coding benchmarks yet still improves a wider task range than the math corpora.
Notably, English-only SFT already transfers to Japanese (MMLU-jp) and Chinese (MMLU-zh) evaluation—see Appendix~\ref{app:xling} for a dedicated cross-lingual analysis. It is also noteworthy that LIMA, a carefully curated dataset for SFT, did not yield substantial performance gains in our controlled setting compared to Alpaca and UltraChat.

Figure~\ref{fig:train-task-accuracy}b plots these relationships separately for each model. Overall tendencies are similar, but there are also considerable differences across models—revealed only because we employed a unified experimental procedure. Some models benefit from almost all training data, whereas others demonstrate minimal gains.

In Figure~\ref{fig:train-task-accuracy}c, we show a correlation matrix of performance gains across different models. As anticipated, models belonging to the same family exhibit high correlations, suggesting that even with additional training, the impact of SFT remains similar within each family. Surprisingly, the language in which a model was initially trained does not appear to substantially affect its overall similarity to others.

Figure~\ref{fig:train-task-accuracy}c also reveals that, in general, the performance structures of the models are quite similar. To examine this more thoroughly, we vertically concatenated the data × benchmark matrices for each model, applied PCA, and then computed the cumulative explained variance ratio (Figure~\ref{fig:train-task-accuracy}d). As shown, about five principal components explain over 90\% of the total variance, indicating a considerable degree of similarity in how different datasets influence SFT outcomes. Nonetheless, certain differences among models persist.

Figure~\ref{fig:train-task-correlation}a, pairwise correlation performance improvements across training datasets, highlights that the similarity or synergy across training datasets varies substantially by model: the same pair of datasets could be complementary in one model but neutral or even conflicting in another. Conversely, Figure~\ref{fig:train-task-correlation}b, pairwise correlation across evaluation tasks, shows a consistency across models, suggesting that tasks requiring similar reasoning skills (e.g., Math tasks) remain closely grouped. A paired t-test on the lower-triangle distributions of Figure~\ref{fig:train-task-correlation}c shows that the correlations across evaluation tasks significantly exceeds that of training datasets ($p<0.01$), confirming that the effects of training datasets is more diverse than evaluation tasks (Figure~\ref{fig:train-task-correlation}d).
Overall, these findings underscore that while some training datasets offer consistent improvements, the degree of benefit often depends on the model. Furthermore, although fine-tuning effects on evaluation tasks are similar across models, those on training datasets are highly model-specific.

\subsection{RQ2. Which Properties of Training Data Matter Most?}
\begin{figure*}[t]
    \centering
    \includegraphics[width=1.0\linewidth]{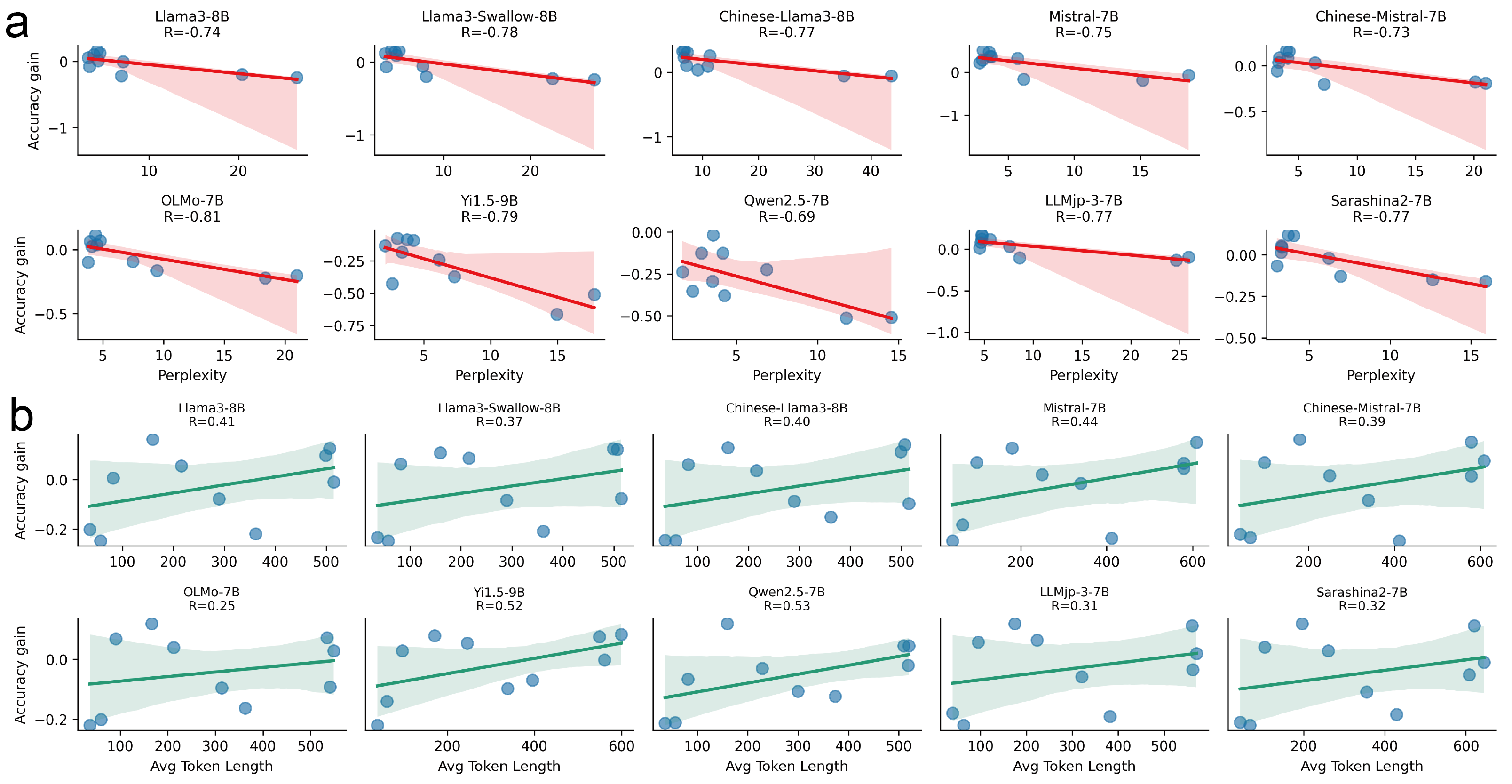}
    \caption{Analysis of training data properties that affect downstream performance. We compare perplexity (\textbf{a}), and  token length (\textbf{b}) with the average performance changes of benchmark tasks for the SFT models, highlighting that lower perplexity is a strong predictor of higher performance.}
    \label{fig:data-properties}
\end{figure*}

Next, we investigate which characteristics of training data most influence performance. Our focus includes perplexity, average token length, and semantic similarity to clarify which factors truly drive effective SFT.

As shown in Figure~\ref{fig:data-properties}a, there is a clear negative correlation in many tasks and models between lower perplexity (w.r.t. the base model) and improved downstream performance. This implies that data lying in a domain or language distribution already ``understood'' by the model can be leveraged more effectively in SFT.

Figure~\ref{fig:data-properties}b reveals a modest correlation between the mean token length of a dataset and downstream performance, suggesting that simply using shorter or longer texts does not strongly drive better results. A prior study has reported that longer texts could be important for performance~\citep{zhao2024long}, and our findings partially support a straightforward link between text length and outcome quality.

Finally, we compare semantic embedding-based similarity between training and evaluation benchmark against performance improvement. Surprisingly, direct semantic similarity is not as strong a predictor as perplexity. Although we observe domain-specific gains (e.g., math data helps on Math tasks, code data helps on coding tasks), a broader trend indicates that linguistic and structural closeness (as reflected in perplexity) may be more decisive than topical resemblance alone. See Appendix~\ref{app:similarity} for the details.

In sum, perplexity relative to the base model emerges as a strong predictor of downstream gains, surpassing factors like token length or broad semantic alignment.
It thus serves as a practical indicator of model–data compatibility for SFT, even when the base model’s pretraining data are unknown.
% It thus serves as a practical indicator of model–data compatibility for SFT.

\subsection{RQ3. Layer-wise weight changes, their relationship to performance, and the effect of SFT on representational dimensionality}
\begin{figure*}[t]
    \centering
    % \fbox{\rule{0.95\linewidth}{4cm}}%
    \includegraphics[width=1.0\linewidth]{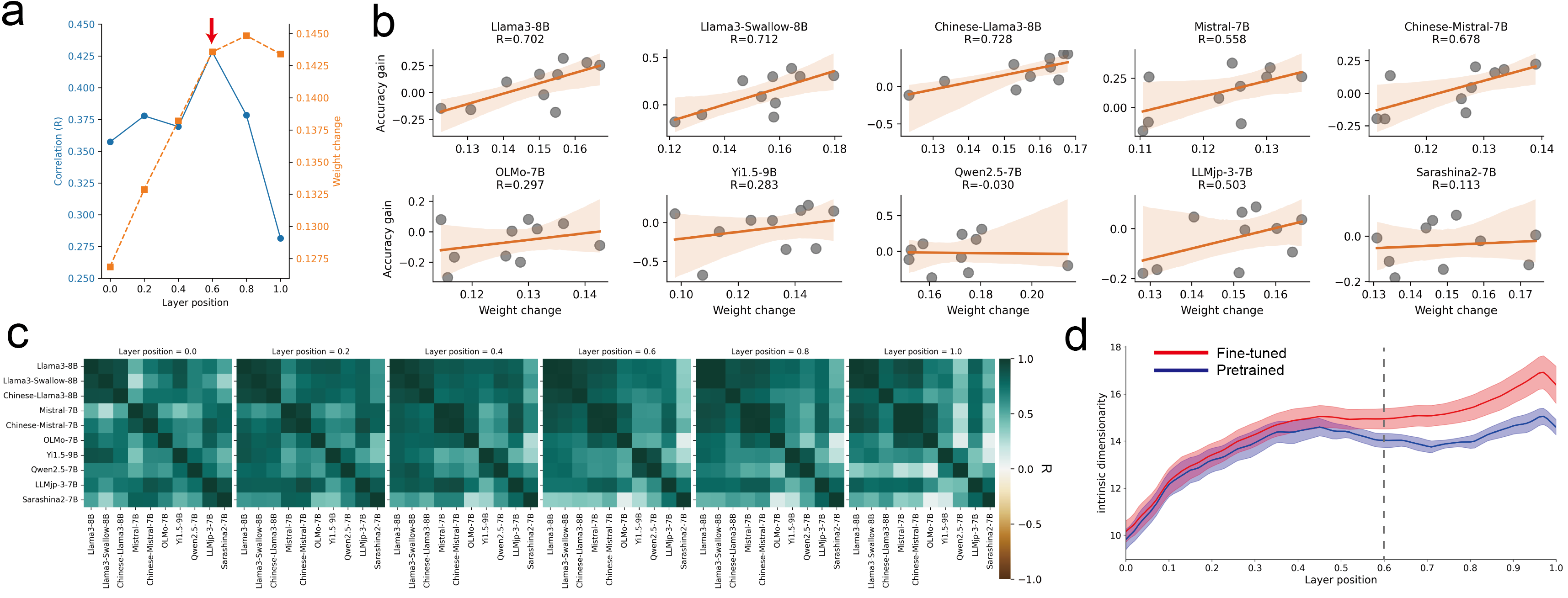}
    \caption{Layer-wise weight changes and their correlations with performance improvements.
\textbf{a} Blue line indicates correlation coefficients between the amount of weight change from the base model and the overall improvement in accuracy, plotted as a function of layer position (0 = input; 1 = output). Compared to early and late layers, the mid-layers (0.6, indicated by red arrow) exhibit the strongest correlation. Orange line indicates the amount of weight change from the base model.
\textbf{b} Focusing on the mid-layer (0.6), examining the relationship between the amount of weight change and accuracy change for each model reveals a robust correlation across all models.
\textbf{c} Correlations calculated across models between weight changes from the base model and those from models trained on specific data. Again, the mid-layers show the strongest model-to-model correlation.
\textbf{d} Intrinsic dimensionality (ID) of training-data embeddings before (blue line) vs. after SFT (red line). The divergence emerges around layer-position = 0.6 (dashed line), suggesting that mid-layer updates expand the representational subspace.
}
    \label{fig:layer-properties}
\end{figure*}

We then explore how model parameters shift during fine-tuning by analyzing layer-wise weight updates across multiple models in detail. Our goal is to identify which layers are most critical in translating SFT into performance gains.

Figure~\ref{fig:layer-properties}a plots two curves: specifically, the blue line is the Pearson correlation between weight-delta magnitude and overall accuracy gain, whereas the orange line shows the raw weight-delta magnitude.
The orange line grows toward upper layers, yet the blue line peaks in the middle, indicating that the largest edits are not the most consequential ones. Rather, we find that the middle layers in particular exhibit the strongest positive correlation with performance gains.

Figure~\ref{fig:layer-properties}b compares the similarity of these layer-wise change patterns across different models. Even though models differ at the architectural level, their mid-layer updates under SFT can follow surprisingly similar trajectories. Still, some model-specific nuances remain.

% Figure~\ref{fig:layer-properties}c extends this idea across models: it correlates, for different layer, the weight-change vector of one model with the corresponding vector of every other model.
Figure~\ref{fig:layer-properties}c quantifies cross-model similarity at each layer. We vectorize the SFT weight change per layer for each model, compute its correlation with the corresponding vectors from all other models, and average the resulting pairwise correlations.
The strongest agreement again lies in the mid-layers, suggesting that SFT enforces a shared instruction-following mechanism across models.

Figure~\ref{fig:layer-properties}d complements the weight-change analysis by quantifying how SFT alters the geometry of the training corpus in embedding space.
For every layer we computed the intrinsic dimensionality (ID) of the sentence-level embeddings produced before and after SFT (methodological details and additional results in Appendix~\ref{app:clustering}).
The difference between the fine-tuned and pretrained ID curves is minimal in the lower half of the network, but from layer-position = 0.6 onward the dimensionality increases sharply and remains elevated through the output layers.
The inflection point coincides with the correlation peaks in Figure~\ref{fig:layer-properties}a, implying that mid-layer updates do more than reduce loss—they actively expand the model’s representational subspace.

Our findings indicate that changes in the mid-layers show the strongest correlation with improved results, suggesting they play a pivotal role in capturing the benefits of SFT.

\subsection{RQ4. Other Factors}
\begin{figure*}[t]
    \centering
    % \fbox{\rule{0.95\linewidth}{5cm}}%
    \includegraphics[width=1.0\linewidth]{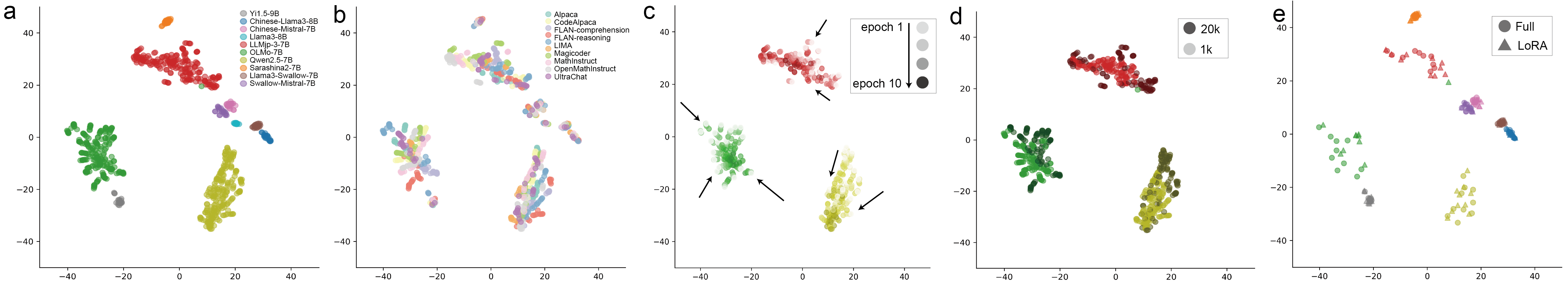}
    \caption{
            t-SNE visualization of log-likelihood vector.  
           \textbf{a}~Colour = model; \textbf{b} colour = training data; \textbf{c} epoch trajectories for three models;  
           \textbf{d} colour = sample size; \textbf{e} shape = tuning method (circle = full, triangle = LoRA).
           }
    \label{fig:other-properties}
\end{figure*}
Finally, we consider additional aspects of SFT, including LoRA versus full-parameter tuning, the effect of sample size, and cross-lingual transfer, each of which may influence final performance.

To disentangle the multiple factors in SFT, we mapped the 757 fine-tuned models—covering 10 base architectures × 10 training datasets and spanning LoRA vs. full-parameter updates, 1–10 training epochs, and sample sizes of 1k or 20k—into a common latent space using log-likelihood-vector projection \citep{oyama2025mapping1000languagemodels}.  
For every model we computed a 1,950-dimensional vector of token-level log-likelihoods by randomly sampling 150 questions from each of the 13 evaluation tasks.
t-SNE then embedded these vectors into two dimensions, giving five complementary views in Fig.\,\ref{fig:other-properties}.
\paragraph{Model families dominate.}  
When points are coloured by \textbf{model} (Fig.\,\ref{fig:other-properties}a) the clusters group almost perfectly by architecture, whereas colouring by \textbf{training data} (Fig.\,\ref{fig:other-properties}b) produces only weak separation. Thus the inductive biases of the base model outweigh the specific SFT corpus in determining the final representation.
\paragraph{Epoch-wise trajectories converge.}  
For the three checkpointed models (Qwen, LLM-jp, OLMo) we plot epochs 1–10 (Fig.\,\ref{fig:other-properties}c).  Irrespective of dataset, trajectories spiral toward a common sub-region, suggesting that SFT gradually aligns the representations toward a shared ``instruction-following'' direction.
\paragraph{Small sample size is often sufficient.}
Colouring by \textbf{training‐set size} separates models trained on 20k samples from those trained on 1k samples.  
The 20k-sample–trained points occupy the outer rim of the manifold more often, whereas the 1k-sample–trained points cluster nearer the core.  
Thus a compact 1k instruction set already supplies sufficient signal for effective instruction-tuning, while scaling up to 20k samples can sometimes pull the representation away from the optimum.
Indeed, our quantitative evaluations showed no consistent accuracy advantage for the 20k-sample models over their 1k-sample counterparts.
\paragraph{LoRA vs.\ full-parameter fine-tuning.}  
Shape-coding full-parameter models as circles and LoRA models as triangles reveals minimal separation; LoRA points are only slightly more peripheral. Quantitatively, full-parameter tuning still excels on reasoning-heavy maths tasks, but LoRA enjoys a small mean advantage on open-ended QA benchmarks.
\paragraph{Cross-lingual transfer persists.}
We also examined the effect of SFT effects on Japanese and Chinese MMLU variants (full results and plots are in Appendix~\ref{app:xling}).  
While we only used English training datasets, performance gains on MMLU are strongly correlated on those of MMLU-jp and MMLU-zh.  
This supports the hypothesis that content overlap between benchmarks, rather than surface-level language similarity, governs cross-lingual transfer in SFT. See Appendix~\ref{app:xling} for the details.

\section{Discussion and Conclusion}
\label{sec:discussion}
We conducted a comprehensive set of SFT experiments involving multiple 7B-scale base models, diverse training datasets, and a wide array of downstream tasks.
Our analysis revealed that, while certain dataset–task synergies are observed consistently across models, their effects can vary greatly depending on the specific model in question.
Notably, perplexity emerged as a particularly robust predictor of SFT success, outperforming both topic similarity and average sequence length.
Perplexity can be shaped by multiple latent properties of both the data and the model. Accordingly, we view it as a practical proxy for compatibility between the model and the data rather than a causal factor. Identifying the causal drivers behind this association is an important direction for future work.

Furthermore, mid-layer weight changes were found to correlate most strongly with performance improvements, indicating that critical adaptations often take place in these layers.
By embedding every model checkpoint into a common latent space, we found that (i)~model architecture exerts a stronger influence than the SFT corpus, (ii)~training epochs drive diverse runs toward a shared instruction-compatible region, (iii)~large instruction sets tend to relocate models toward the periphery—often reducing accuracy relative to smaller sets—and (iv)~LoRA trajectories almost coincide with full-parameter ones, diverging only slightly on the periphery; this mirrors the small but systematic trade-off we observed between knowledge-heavy tasks (full-parameter advantaged) and open-ended QA (LoRA advantaged).

Contrary to the typical assumption that a dataset closely resembling the target task is best, we find data with a lower perplexity (where the model requires minimal additional learning or unlearning) generally yields more robust improvements. Additionally, our observations of code data helping math tasks suggest significant cross-domain transfer beyond simple topic alignment.

Discovering the importance of mid-layer changes could reshape fine-tuning strategies. Updating the mid-layers, or monitoring them closely, could provide more efficient or interpretable SFT. Observing common mid-layer change patterns across models suggests a shared mechanism for task-related knowledge acquisition.
\section*{Limitations}
In this study, we conducted comprehensive fine-tuning experiments on pre-trained models in the 7-9B parameter class. Due to limited computational resources, we were unable to extend our verification to larger models (e.g., 70B, 175B, or MoE architectures). It is unclear whether the findings obtained in this research can be generalized to these larger models. 

We emphasize the transparency and reproducibility of our analysis, and therefore, we trained our models using only open-access training datasets. Compared to existing publicly available English corpora, there are still limited multilingual instruction-tuning datasets. Consequently, our study used only English training datasets. A comprehensive investigation into cross-lingual knowledge transfer and its differing effects remains a subject for future work. We use about 10 popular training datasets for SFT, possibly limiting generality for highly specialized tasks or broader multilingual corpora.

While perplexity proved insightful, it can fluctuate based on tokenizer design and base training distributions, indicating a need for more nuanced measures.
\section*{Ethical Considerations}
This work uses only publicly available and properly licensed datasets and base models. Their licenses permit research use and redistribution. All datasets and models were used in accordance with their intended research purposes, and our released models will maintain this intended use.

We did not collect any new data. While we did not manually inspect all samples, we acknowledge the possibility of residual personally identifiable or harmful content in the original datasets and rely on the original curators’ filtering processes.

We will release over 1,000 fine-tuned models as part of this study. While we do not anticipate major risks, we acknowledge the potential for misuse—such as generating harmful or misleading content. To mitigate this, all released models will include a responsible use clause and detailed model cards describing limitations. We encourage responsible use.

We used AI tools to assist in writing training and evaluation scripts, and to support basic analysis tasks such as summarizing experimental results.
\section*{Acknowledgements}
\label{sec:acknowledgements}
We used the ``mdx: a platform for building data-empowered society''~\citep{mdx} for our experiments and analyses.

\bibliographystyle{acl_natbib}

% \bibliography{custom}

\appendix
\renewcommand{\thefigure}{A.\arabic{figure}}
\setcounter{figure}{0}

%========================================%
\section{Description of Base Models}
\label{sec: description of base models}

% テーブル+詳細
\begin{table*}[]
\begin{tabular}{@{}llll@{}}
\toprule
\textbf{Language}         & \textbf{Model Name (Params)} & \textbf{Repository (Hugging Face)}     & \textbf{Context Length} \\ \midrule
\multirow{4}{*}{English}  & OLMo (7B)                    & allenai/OLMo-7B-hf                    & 2048                    \\
                          & Llama3 (8B)                  & meta-llama/Meta-Llama-3-8B            & 8192                    \\
                          & Mistral (7B)                 & mistralai/Mistral-7B-v0.1             & 32768                   \\
                          & Gemma2 (9B)                  & google/gemma-2-9b                     & 8192                    \\ \midrule
\multirow{4}{*}{Chinese}  & Qwen2.5 (7B)                 & Qwen/Qwen2.5-7B                       & 131072                  \\
                          & Chinese-Llama3 (8B)          & hfl/llama-3-chinese-8b                & 8192                    \\
                          & Chinese-Mistral (7B)         & itpossible/Chinese-Mistral-7B-v0.1    & 32768                   \\
                          & Yi1.5 (9B)                   & 01-ai/Yi-1.5-9B                       & 4096                    \\ \midrule
\multirow{4}{*}{Japanese} & LLMjp-3 (7B)                 & llm-jp/llm-jp-3-7.2b                  & 4096                    \\
                          & Llama3-Swallow (8B)          & tokyotech-llm/Llama-3-Swallow-8B-v0.1 & 8192                    \\
                          & Swallow-Mistral (7B)         & tokyotech-llm/Swallow-MS-7b-v0.1      & 4096                    \\
                          & Sarashina2 (7B)              & sbintuitions/sarashina2-7b            & 4096                    \\ \bottomrule
\end{tabular}
\caption{Overview of the 12 base models employed for SFT experiments. The table summarizes their parameter sizes, primary training language, and maximum supported context lengths.}
\label{tab:basemodel}
\end{table*}

\textbf{OLMo}~\citep{Groeneveld2023OLMo} is developed by \emph{Allen Institute for AI}. An English-centric, 7B-parameter decoder model pre-trained on a carefully filtered mix of web pages, books, and code (totaling 2.5 trillion tokens). Flash-Attention 2 support was added in later versions, enabling fast, memory-efficient inference. Model-card results show competitive GSM8K and MMLU scores, rivaling some 10B-class models.

\textbf{Llama3}~\citep{dubey2024llama} is developed by \emph{Meta AI}. An 8B English model trained on multi-trillion-token mixed-domain data with a byte-level BPE tokenizer and scaled RoPE. Safety alignment combines RLHF and rejection sampling. Delivers strong, well-rounded performance across reasoning, code, and chat benchmarks.

\textbf{Mistral}~\citep{jiang2023mistral7b} is developed by \emph{Mistral AI}. An English 7B model whose pre-training corpus mixes web, academic text, and code. Grouped-query and sliding-window attention enable very long-sequence processing while retaining high speed. Matches or exceeds Llama-2-13B on many English tasks.

\textbf{Gemma2}~\citep{gemmateam2024gemma2improvingopen} is developed by \emph{Google DeepMind}. An English 9B model trained on a large quality-filtered corpus and enhanced with internal architectural refinements such as improved normalization and position encoding, building on modern Transformer techniques. Public reports show it surpasses most open 7–13B baselines on language-understanding leaderboards.

\textbf{Qwen2.5}~\citep{qwen2} is developed by \emph{Alibaba’s Qwen team}. A Chinese–English bilingual 7B model further pre-trained on high-quality proprietary Chinese data. RoPE extrapolation enables extremely long inputs. The model card provides agent-style prompting templates and strong results on tool use and code generation.

\textbf{Chinese-Llama3}~\citep{chinese-llama-alpaca} is developed by \emph{Harbin NLP (HFL)}. An 8B Chinese model obtained by continual pre-training of Llama-3 on an extensive Chinese corpus with vocabulary augmentation. Significantly boosts Chinese QA and CMMLU scores over the original Llama-3.

\textbf{Chinese-Mistral}~\citep{hsu2024breeze7btechnicalreport} is developed by \emph{itpossible}. A 7B Chinese variant of Mistral-v0.1, additionally trained on Chinese Wikipedia, news, and conversation data. Improves cross-lingual performance on Chinese benchmarks while preserving the original architecture.

\textbf{Yi1.5}~\citep{ai2025yiopenfoundationmodels} is developed by \emph{01.AI}. A 9B multilingual model (Chinese + English focus) based on the original Yi model trained on 3.1 trillion tokens, with an additional 500 billion tokens used for continual pretraining, including substantial code and low-resource-language data. Shows solid zero-shot transfer to many Asian and European languages as well as code-related tasks.

\textbf{LLMjp-3}~\citep{llmjp2024llmjpcrossorganizationalprojectresearch} is developed by \emph{LLM-jp}. A 7.2B Japanese-centric model built from scratch on a 2.1 trillion token multilingual corpus, predominantly composed of Japanese web, book, and dialogue texts, along with a smaller portion of English and other languages. Public experiments indicate it surpasses Llama-2-13B on Japanese QA and summarization.

\textbf{Llama3-Swallow}~\citep{fujii2024continualpretrainingcrosslingualllm} is developed by \emph{TokyoTech LLM Group}. An 8B Japanese model produced by continual pre-training of Llama-3-8B on large Japanese corpora plus vocabulary extension. Reports notable gains for Japanese NER and academic-paper summarization.

\textbf{Swallow-Mistral}~\citep{fujii2024continualpretrainingcrosslingualllm} is developed by \emph{TokyoTech LLM Group}. A 7B Japanese follow-up to Mistral-7B with memory-footprint optimizations. Excels at Japanese dialogue and technical writing according to model-card evaluations.

\textbf{Sarashina-2}\footnote{https://huggingface.co/sbintuitions/sarashina2-7b} is developed by \emph{sbintuitions}. A 7B Japanese Llama derivative further trained on Japanese text and code. Distributed with LoRA adapters, making domain-specific fine-tuning straightforward.

%========================================%

\section{Description of Training Datasets}
\label{sec: description of training datasets}
% テーブル+詳細
% outputにnoneがあるサンプルはスキップする処理をしたこと（flan_knowledge, flan_comprehension）
% ultrachatは最初のターンを使用したこと
% ultrachatは本家（https://huggingface.co/datasets/stingning/ultrachat）でないものを使用している
% 平均文長には全て1kデータを使用しているが、20kデータの場合とそこまで変わらなかった。また、入出力の両方を合わせた数字になっている。

\begin{table*}[]
\begin{tabular}{lllll}
\hline
\textbf{Category}            & \textbf{Dataset}   & \textbf{Repository (Hugging Face)}     & \textbf{Samples} & \textbf{Lengths} \\ \hline
\multirow{3}{*}{General}     & Alpaca             & yahma/alpaca-cleaned                  & 51,760           & 122.19                                \\
                             & LIMA               & GAIR/lima                             & 1,330            & 391.97                                \\
                             & Ultrachat          & HuggingFaceH4/ultrachat\_200k         & 773,913          & 397.45                                \\ \hline
\multirow{2}{*}{Coding}      & CodeAlpaca         & sahil2801/CodeAlpaca-20k              & 20,022           & 44.36                                \\
                             & Magicoder          & ise-uiuc/Magicoder-Evol-Instruct-110K & 111,183          & 300.04                                \\ \hline
\multirow{2}{*}{Math}        & OpenMathInstruct  & nvidia/OpenMathInstruct-1             & 6,078,712        & 140.43                                \\
                             & MathInstruct       & TIGER-Lab/MathInstruct                & 262,039          & 125.36                                \\ \hline
\multirow{3}{*}{\begin{tabular}[c]{@{}l@{}}Classic\\ NLP\end{tabular}} & FLAN Knowledge     & Open-Orca/FLAN                        & 226,575          & 21.69                                \\
                             & FLAN Reasoning     & Open-Orca/FLAN                        & 92,770           & 41.74                                \\
                             & FLAN Comprehension & Open-Orca/FLAN                        & 208,605          & 262.31                                \\ \hline
\end{tabular}
\caption{Repository is the original source of the data used and Samples represents its total number of samples. Lengths indicates the average number of words in each data at 1k sample pre-processing.}
\label{tab:training-data}
\end{table*}

\textbf{Alpaca}~\citep{alpaca} is a 52k-example English corpus obtained by filtering the original Stanford Alpaca to remove hallucinating prompts, merged instructions, empty outputs, and other defects. The resulting instruction/input/output triples serve as a cleaner general-purpose starting point for instruction tuning.

\textbf{LIMA}~\citep{zhou2024lima} is a compact set of 1000 prompt–response pairs—750 mined from Stack Exchange, wikiHow, and r/WritingPrompts plus 250 author-written items—selected for diversity and a consistent assistant style. It probes how well a strong language model can be aligned with minimal but high-quality supervision.

\textbf{UltraChat}~\citep{ding2023enhancing} is a 774k multi-turn English dialogue corpus synthesized by two ChatGPT-Turbo agents.  
We use a reformatted version of the original release \footnote{\url{https://huggingface.co/datasets/stingning/ultrachat}}.  
In our preprocessing pipeline, we extract only the initial user prompt and the first assistant reply as each training sample.  

\textbf{CodeAlpaca 20k}~\citep{codealpaca} is a collection of 20k English programming instructions generated with the Self-Instruct pipeline using text-davinci-003. About 40\% of the samples include an input field, and the schema mirrors Alpaca but focuses exclusively on code generation and editing.\footnote{\url{https://github.com/sahil280114/codealpaca}}

\textbf{Magicoder}~\citep{wei2024magicoder} contains 111k licence-clean code-centric instructions obtained by de-contaminating the Evol-CodeAlpaca corpus. Every example is a single-turn instruction→response pair, offering a larger companion to CodeAlpaca.

\textbf{OpenMathInstruct}~\citep{toshniwal2024openmathinstruct} is a 1.8M mathematics corpus whose step-by-step solutions were generated with Mixtral-8×7B and a Python interpreter, then automatically validated.

\textbf{MathInstruct}~\citep{yue2023mammoth} aggregates 262k math-reasoning problems from 13 sources and augments them with both chain-of-thought and program-of-thought rationales, supplying lightweight yet generalizable coverage for mathematical fine-tuning.

\textbf{FLAN Collection}~\citep{weifinetuned} is the remix file flan2021\_zsnoopt\_submix\_data.json. Specifically, it corresponds to the FLAN‑2021 sub‑mix and employs the zero‑shot, no‑options template variant (i.e., prompts contain only the instruction without in‑context examples or candidate options). We follow the taxonomy of \citet{dubey2024llama,2023opencompass} and split the data into three thematic subsets.

\textbf{FLAN Knowledge} uses BoolQ (bool\_q:1.0.0), NaturalQuestions (natural\_questions\_open:1.0.0), and TriviaQA (trivia\_qa/rc:1.1.0). Samples whose output field is "none" are discarded.
% bool\_q:1.0.0, 8734 \citep{clark2019boolq}
% natural\_questions\_open:1.0.0, 82295 \citep{47761}
% trivia\_qa/rc:1.1.0, 81142 \citep{joshi-etal-2017-triviaqa}

\textbf{FLAN Reasoning} combines ARC-Easy (ai2\_arc/ARC-Easy:1.0.0), ARC-Challenge (ai2\_arc/ARC-Challenge:1.0.0), HellaSwag (hellaswag:1.1.0), WinoGrande (winogrande:1.1.0), and PIQA (piqa:1.0.0).
% ai2\_arc/ARC-Easy:1.0.0, 1891 \citep{allenai:arc}
% ai2\_arc/ARC-Challenge:1.0.0, 850 \citep{allenai:arc}
% hellaswag:1.1.0, 37295 \citep{zellers2019hellaswag}
% winogrande:1.1.0, 37731 \citep{sakaguchi2019winogrande}
% piqa:1.0.0, 15003 \citep{Bisk2020}

\textbf{FLAN Comprehension} contains QuAC (quac:1.0.0) and SQuAD v2.0 (squad/v2.0:3.0.0). Samples with an output of `none` are omitted.
% quac:1.0.0, 78270 \citep{choi-etal-2018-quac}
% squad/v2.0:3.0.0, 121601 \citep{rajpurkar-etal-2018-know}

%========================================%

\section{Preliminary Experiments}
\label{sec: preliminary experiments}

% FLANはこの時点では1つにまとまっていたことの説明が必要
% 結果？
% 2kで試したのは（本実験と同じ設定なので）予備実験には入らない気がする？
In our main experiments, we conduct SFT using various base models and diverse training datasets. To ensure valid and reliable results across different configurations, it is crucial to select appropriate hyperparameters. Therefore, we conducted preliminary experiments aimed at determining suitable hyperparameters.

These preliminary experiments were carried out under the following conditions. We employed the Llama3-8B model and utilized six different datasets, each comprising approximately 1,000 samples: Magicoder, LIMA, Code Alpaca, FLAN, Openmath, and Alpaca.

% We examined several hyperparameter settings: learning rate = \texttt{\{2e-7, 1e-6, 2e-6, 1e-5, 2e-5, 1e-4\}}, batch size = \texttt{\{32, 64, 128, 256\}}, weight decay = \texttt{\{0, 0.1\}}, training method = \texttt{\{LoRA, full-parameter tuning\}}.
We examined several hyperparameter settings: learning rate = \{2e-7, 1e-6, 2e-6, 1e-5, 2e-5, 1e-4\}, batch size = \{32, 64, 128, 256\}, weight decay = \{0, 0.1\}, training method = \{LoRA, full-parameter tuning\}.

This combination of hyperparameters resulted in 96 unique experimental conditions. For each condition, we trained for 10 epochs and saved a checkpoint after every epoch (epochs 1 to 10). We treated these per-epoch checkpoints as distinct candidate models, yielding 960 models per dataset. Given that we utilized six datasets, the total number of trained models reached 5,760, consuming at least 23,040 GPU-hours.

For evaluation purposes, we utilized two benchmarks: MMLU and MT-bench, ensuring comprehensive performance assessment across diverse tasks.

\section{Description of Training Settings}
\label{sec: description of training settings}
This section summarizes the training configurations, computational cost, and other implementation details used in our supervised fine-tuning experiments.

Out of a total of 1,070 training runs, 1,059 models were successfully trained. Training failed in 11 cases, all related to out-of-memory (OOM) errors involving the Gemma model trained on the Magicoder dataset. Specifically, one failure occurred in a single-dataset setting, while the remaining ten failures arose during the All Dataset setting, where checkpoints were saved at every epoch (resulting in ten distinct training jobs).

We used separate hyperparameter settings for full-parameter fine-tuning and LoRA. Full-parameter fine-tuning was conducted with a learning rate of \(1.0 \times 10^{-5}\), batch size of 32, weight decay of 0.0, and 10 training epochs. For LoRA, we used a learning rate of \(2.0 \times 10^{-6}\), batch size of 128, weight decay of 0.0, and the same number of epochs. These values were determined based on preliminary grid-search experiments.

The computational time for fine-tuning on 1k samples varied depending on the model, batch size, and training method, but on average, each run took approximately 30 minutes. To accelerate training, we employed Flash Attention 2~\citep{dao2023flashattention2fasterattentionbetter} and DeepSpeed~\citep{deepspeed2020} for all models. Training these 1,059 runs required at least 530 GPU-hours.

To investigate the impact of individual datasets, we conducted a dataset ablation study using three representative models: OLMo, Qwen, and LLM-jp. In this setting, we trained models on nine datasets at a time, excluding one dataset in each run (i.e., leave-one-out strategy). This allowed us to observe how the absence of specific datasets affected downstream performance. The ablation experiments were performed under the same conditions as regular 1k-sample training, using both full-parameter and LoRA-based fine-tuning.

As noted in Section~\ref{sec: Training Datasets}, all datasets were preprocessed under consistent conditions. During training, we formatted all samples using a standardized instruction-response template:
\begin{verbatim}
###Question: {instruction}
###Answer: {response}
\end{verbatim}

\section{Description of the Evaluation Dataset}
\label{sec: description of the dataset}
This appendix provides an overview of the datasets used for evaluation. The test cases and evaluation settings follow the format provided by OpenCompass.
% Opencompassのベンチマークの説明文をそのまま抜粋

%MATH
\textbf{MATH} \citep{MATH} consists of 12,500 problems from high school math competitions. Each problem in MATH has a full step-by-step solution and models are tasked with generating tokens to construct the final answer.

%GSM8k
\textbf{GSM8K} \citep{GSM8k} is a dataset of 8,500 high quality linguistically diverse grade school math word problems created by human problem writers. Compared to MATH, the problems are easier and include basic knowledge questions, such as asking for the number of days in a week.

%HumanEval
\textbf{HumanEval} \citep{HumanEval} consists of 164 hand written programming problems. It assesses language comprehension, reasoning, algorithms, and simple mathematics.

%MBPP
\textbf{MBPP} \citep{MBPP} consists of 974 programming tasks, designed to be solvable by entry-level programmers. The problems range from basic computations to those requiring external mathematical knowledge.

%BoolQ
\textbf{BoolQ} \citep{clark2019boolq} is a question answering dataset for yes/no questions containing 15942 examples. The questions are real user queries—unprompted and written without knowing the answers—making them more inferential and challenging than synthetic datasets.

%NQ
\textbf{NaturalQuestion} \citep{NQ} consists of over 300,000 questions. This corpus features questions posed by actual users and challenges QA systems to read and understand a full Wikipedia article, which might or might not include the answer.

%MMLU
\textbf{MMLU} \citep{MMLU1, MMLU2} consists of multiple-choice question-answer pairs divided into 57 subjects spanning STEM fields, the humanities, social sciences, and beyond. The questions vary in complexity, from elementary to expert-level, and assess both factual knowledge and reasoning skills.  

%CMMLU
\textbf{MMLU-zh} \citep{CMMLU} contains 11,528 multiple-choice questions across 67 diverse subjects, including STEM, humanities, social sciences, and China-specific topics (e.g., Chinese law, traditional medicine, and ancient Chinese). The dataset is specifically constructed to reflect the linguistic and cultural nuances of Chinese, with many questions that are not easily translatable from English benchmarks like MMLU.

%JMMLU
\textbf{MMLU-jp} We evaluated the models’ Japanese generation ability using the Japanese-translated version of the multilingual MMLU test set \footnote{\url{https://huggingface.co/datasets/openai/MMMLU}}. While the content of the questions remains the same as the original MMLU, both the questions and answers are presented in Japanese.

%TruthfulQA
\textbf{TruthfulQA} \citep{TruthfulQA} comprises 817 questions that span 38 categories, including health, law, finance and politics. The questions are carefully crafted to trigger imitative falsehoods—answers that are commonly believed but factually incorrect.

%MT-Bench
\textbf{MTBench} \citep{MTBench} is a benchmark designed to evaluate a model’s instruction-following capabilities in a multi-turn dialogue format, consisting of 80 two-turn question sets. We conducted evaluations using the LLM-as-a-Judge framework, employing gpt-4o-2024-08-06 as the evaluator LLM.

%Alpaca Eval v2
\textbf{AlpacaEval v2} \citep{AlpacaEvalv2}
AlpacaEval v2 is an instruction-following benchmark consisting of 805 questions, created by integrating existing benchmarks and incorporating insights from real user interactions. We conducted pairwise evaluations by comparing the responses of our fine-tuned models against those of GPT-4-Turbo, and report the win rate as the evaluation metric. We used gpt-4o-2024-07-18 as the evaluator LLM.

All models evaluated in this experiment used the same prompt across all benchmarks. Therefore, it should be noted that the scores on downstream tasks may differ from those reported in technical reports, as the pre-trained models used a prompt template that differs from the one originally provided.
Among the trained models, the following models failed to produce results for certain tasks:
\begin{itemize}
    \item \textbf{Swallow-Mistral-7B}: All 41 trained models encountered out-of-memory errors across all tasks.
    \item \textbf{Mistral-7B}: The 20 LoRA-tuned models encountered out-of-memory errors on few-shot tasks.
    \item \textbf{Gemma2-9B}: For the models trained with all data using LoRA (from epoch 1 to epoch 4), responses that made evaluation with Alpaca Eval v2 impossible (extremely long, repetitive outputs) were generated. As a result, the win rates were recorded as \texttt{NaN}.
\end{itemize}
The completion of the evaluation process required approximately 16,700 GPU-hours, including the results from models and downstream tasks not described in the paper.

\begin{table*}[t]
    % \small
    \centering
    \captionsetup{justification=centering}
    \label{table:evaluationbenchmark}
    \begin{tabular}{llll} \toprule
        \textbf{Category} & \textbf{Dataset} & \textbf{queries} & \textbf{metric} \\ \hline
        \multirow{2}{*}{Math} & MATH & 5000 & Exact Match Accuracy\\ %open compassは回答の正規化処理とparsingがかなり細かく設定されているので、Lenient  matchingともいえる
                              & GSM8K & 1319 & Exact Match Accuracy\\ \hline %Lenient matching
        \multirow{2}{*}{Coding} & HumanEval & 164 & pass@1\\ %成功数/問題数
                                 & MBPP & 257 & pass@1\\ %成功数/問題数
                                 \hline
        \multirow{3}{*}{Knowledge} & BoolQ & 3270 & Exact Match Accuracy \\%parsingして得られた選択肢がgroud truthと一致していれば正解 
                                    & NaturalQuestions & 3610 & Lenient Matching Accuracy\\%grount truthがparsingした出力文字列中に含まれていれば正解
                                    & TruthfulQA & 817 & BLEU Accuracy\\%reference answerと出力のblueが0.7以上であれば正解としたときの割合
                                    \hline
        \multirow{3}{*}{Examination} & MMLU & 14042 & Exact Match Accuracy\\%parsingして得られた選択肢がgroud truthと一致していれば正解 
                                     & MMLU-zh & 11582 & Exact Match Accuracy\\%parsingして得られた選択肢がgroud truthと一致していれば正解 
                                     & MMLU-jp & 14042 & Exact Match Accuracy\\%parsingして得られた選択肢がgroud truthと一致していれば正解 
                                     \hline
        \multirow{2}{*}{Instruction-following} & MT Bench & 160 & Total Score\\ %10点満点
                                    & Alpaca Eval v2 & 805 & Win rate \\ %gpt-4-turboに対する勝率
                                    \bottomrule
    \end{tabular}
    \caption{The list of downstream tasks used to evaluate the fine-tuned models is shown. 
    All tasks are supported by the OpenCompass library, and the evaluation metrics are consistent with those used in OpenCompass.}
\end{table*}

%========================================%

\section{Additional Results: Cross-lingual Transfer}
\label{app:xling}
\begin{figure}[H]
    \centering
    \includegraphics[width=\linewidth]{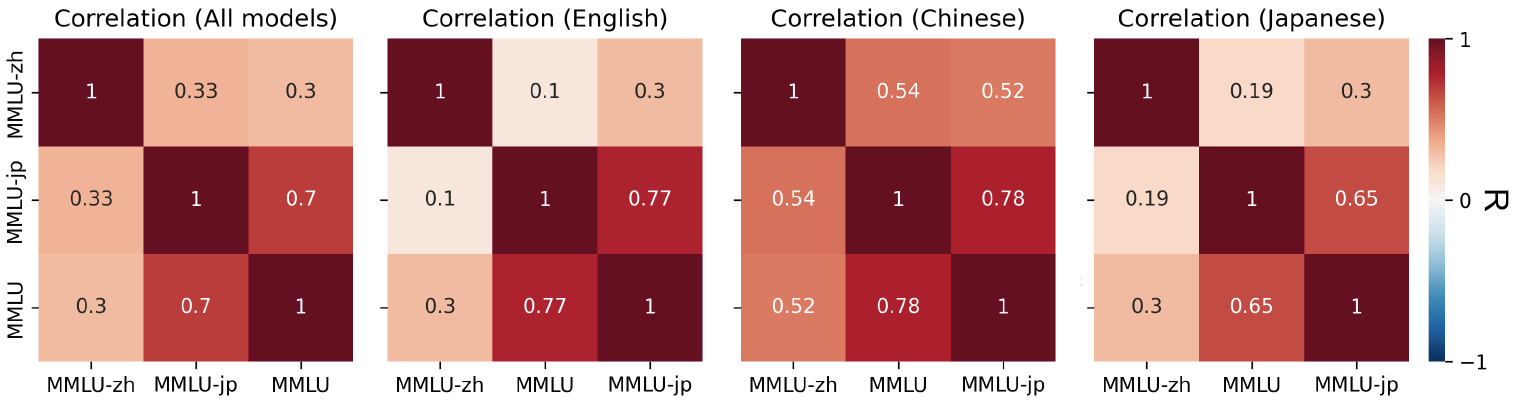}
    \caption{Correlation coefficients of performance gain across models, focusing on MMLU, MMLU-jp and MMLU-zh, split by training languages for the models.}
    \label{fig:cross-lang}
\end{figure}

\noindent
We group models by their pre-training language (English, Chinese, Japanese) and compute pairwise Pearson correlations between MMLU-family scores across English, Chinese, and Japanese test sets (Figure~\ref{fig:cross-lang}).
All language pairs show strong positive correlations: substantial zero-shot transfer even though every SFT run used only English data.
Evaluating SFT conducted in multiple languages remains an open avenue for future work.

\section{Additional results: semantic similarity between training dataset and evaluation tasks}
\label{app:similarity}
To calculate semantic similarity between training datasets and evaluation tasks, we computed BERTScore F1 between every training-dataset and evaluation-task pair using a pretrained BERT-base model. Correlating these scores with the average SFT performance gains yielded only a small, non-significant positive relationship (Pearson's R = 0.112, $P > 0.05$). Hence, semantic closeness—as captured by BERTScore—offers little predictive value for fine-tuning benefit.

%========================================%
\section{Analyzing Hidden Representation Shift}
\label{app:clustering}
We analyze the impact of fine-tuning on the hidden representations of LLMs. Previous studies have shown that task-specific information is encoded in the intermediate layers, and predictions are adjusted toward task-specific representations in the subsequent layers \citep{zhao-etal-2024-layer}. We analyzed the global and local structural changes in the representation space by performing clustering analysis on the hidden representations of the training dataset.

\begin{figure*}[t]
    \centering
    % \fbox{\rule{0.95\linewidth}{4cm}}%
    \begin{subfigure}[b]{0.95\textwidth}
    \includegraphics[width=\linewidth]{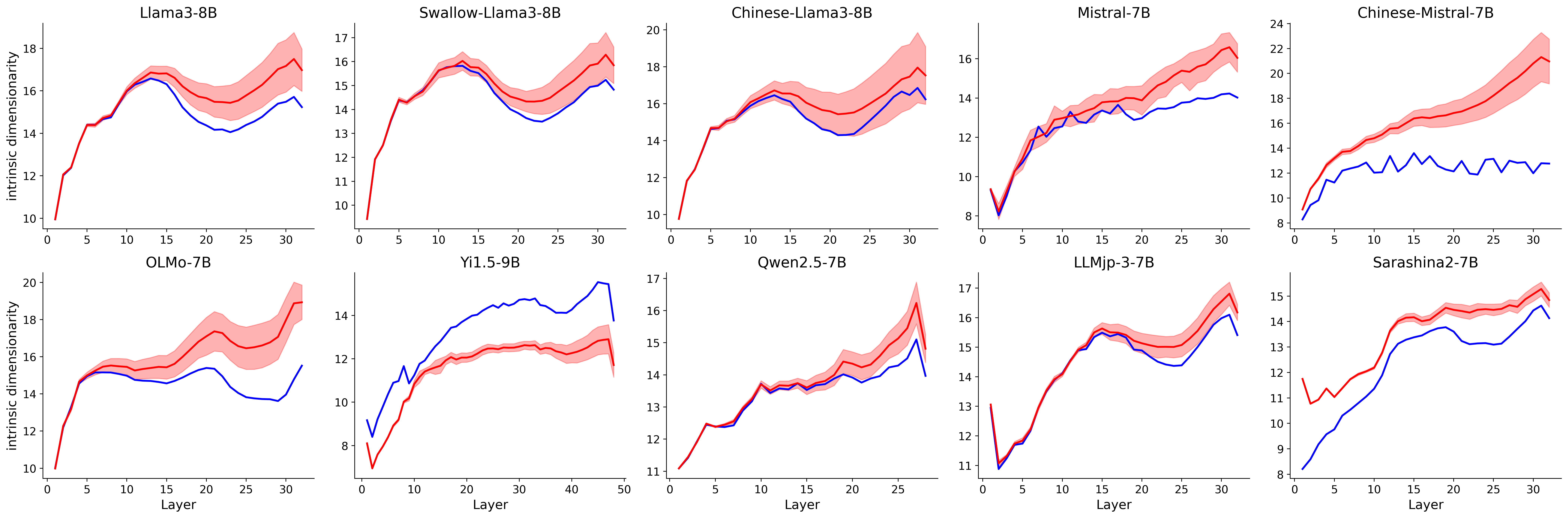}
    \caption{Intrinsic Dimensionality across layers}
    \label{fig:ID_all}
    \end{subfigure}
    \\[1ex]
\begin{subfigure}[b]{0.95\textwidth}
    % \centering
    % \fbox{\rule{0.95\linewidth}{4cm}}%
    \includegraphics[width=\linewidth]{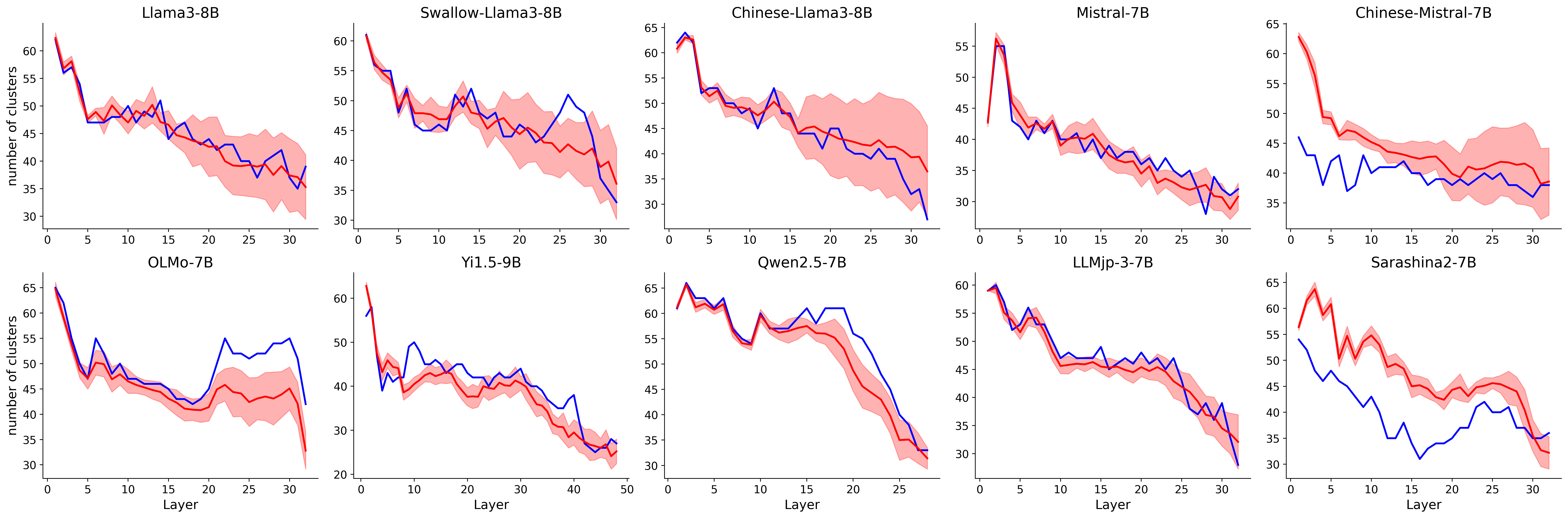}
    \caption{Number of clusters across layers}
    \label{fig:number_of_cluster}
    \end{subfigure}
    \\[1ex]
\begin{subfigure}[b]{0.95\textwidth}
    % \centering
    % \fbox{\rule{0.95\linewidth}{4cm}}%
    \includegraphics[width=\linewidth]{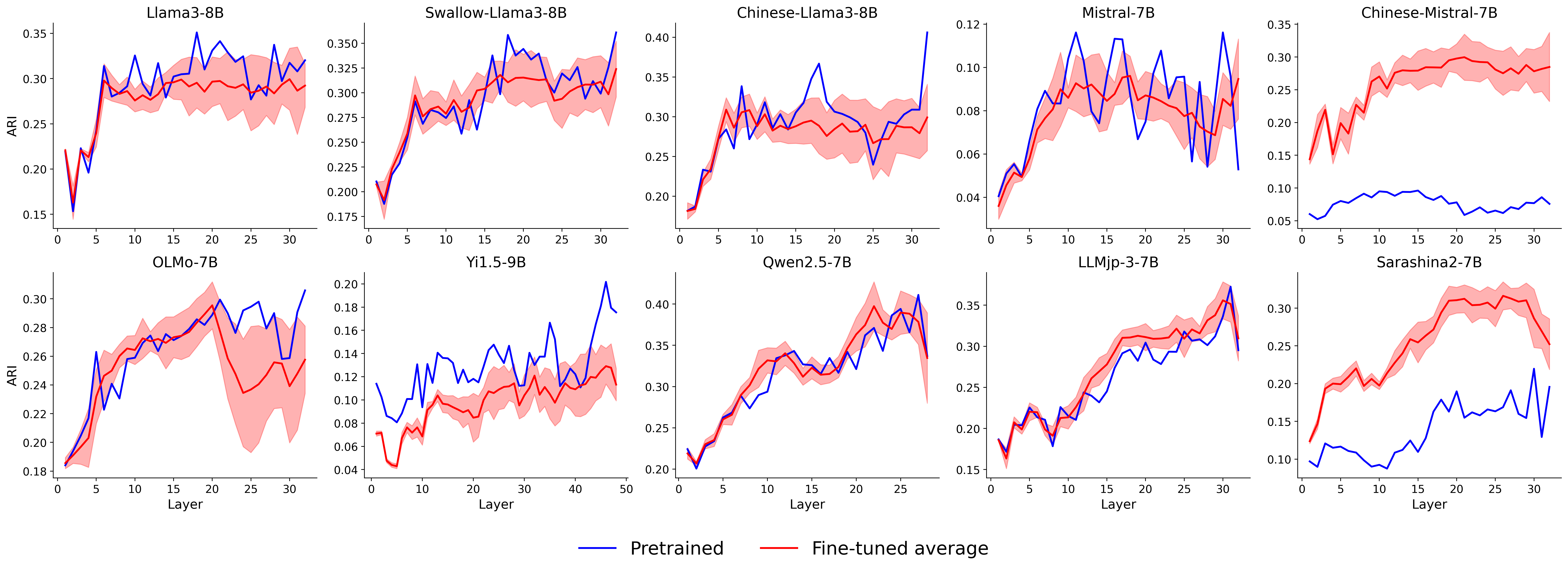}
    \caption{Adjusted Rand Index across layers}
    \label{fig:ARI_all}
\end{subfigure}
\caption{
(a) \textbf{Intrinsic dimensions (ID) per layer of the sentence embeddings from the training dataset.} The blue lines represent the ID of each pretrained model. The red lines indicate the average ID of the models fine-tuned on a single training dataset for each pretrained model. Except for Yi1.5-9B, all models show an increase in ID due to fine-tuning, with the difference becoming apparent from the middle layers onward.\\
(b) \textbf{Number of clusters in the hidden representation space.} In many models, the final layer has the fewest number of clusters. Furthermore, fine-tuning reduces the number of clusters, showing a strong negative correlation with ID.\\
(c) \textbf{Adjusted Rand Index (ARI) of the density based clustering.} The values and trends vary significantly across models and layers. This suggests that the local structural arrangement of the training dataset is highly sensitive to the influence of the model and the dataset.
}
\end{figure*}

\subsection{Methods}
\textbf{LLMs and token representations analyzed.}

We analyzed the hidden representations of a total of 110 models. This includes 10 base models that were primarily used in our evaluation. In addition to the base models, we also analyze the fine-tuned models that were trained on 1k-example subsets from 10 training datasets. This represents the simplest training setup, and analyzing models trained under different settings remains a topic for future work. To examine the effects of fine-tuning on both in-distribution and out-of-distribution data, we used a collection of 1k-example subsets from the training datasets, totaling $N$ = 9974 examples. Following prior work \citep{doimo2024representationlandscapefewshotlearning}, we extracted the embedding of the final token in the prompt of each training example—formatted as \texttt{\#\#\#Question: \{instruction\}}—from the outputs of all Transformer blocks. The final token is expected to encode contextual information accumulated from the preceding input, and its embedding is likely to vary across fine-tuned models. 
Finally, for each model, we obtain an L-layer, d-dimensional embedding space for N examples.

\noindent \textbf{Representation quality measures.}

To quantitatively evaluate the properties of text embeddings, we apply the Advanced Density Peaks (ADP) algorithm \citep{DERRICO2021476}, a density-based clustering method. 
% Compared to Density-based spatial clustering of applications with noise \citep{ester1996density}, another density-based clustering technique, ADP estimates the intrinsic dimensionality and adjusts the density computation accordingly, enabling it to more reliably extract cluster structures even in high-dimensional spaces.
The algorithm first estimates the intrinsic dimensionality (ID) using the Gride \citep{denti2022generalized}. ID reflects how many parameters are needed to describe the data manifold. Based on this estimated dimensionality, it detects local density peaks using neighborhood-based criteria, and retains only statistically significant peaks via a t-test to form $s$ clusters. To evaluate the properties of the estimated clusters, we use the Adjusted Rand Index (ARI) \citep{hubert1985comparing, steinley2004properties}.
ARI is computed by the following formula.
\begin{equation}
\mathrm{ARI} = 
\frac{
\sum_{ij} \binom{N_{ij}}{2}
- 
\frac{
\sum_i \binom{a_i}{2} \sum_j \binom{b_j}{2}
}{
\binom{N}{2}
}
}{
\frac{1}{2} \left[
\sum_i \binom{a_i}{2} + \sum_j \binom{b_j}{2}
\right]
-
\frac{
\sum_i \binom{a_i}{2} \sum_j \binom{b_j}{2}
}{
\binom{N}{2}
}
}
\end{equation}
where $a_i$ represents the number of samples contained in the cluster $A_i$ ($1 \leq i \leq s$) and $b_j$ represents the number of samples belonging to the dataset label $B_j$ ($1 \leq j \leq 10$).
ARI measures how well the found clusters match the ground-truth labels, adjusting for chance grouping. It counts pairwise agreements between the clustering and true labels. 1.0 denotes perfect recovery of true classes by clusters, 0 indicates random alignment, and negative values imply worse-than-random clustering. 
% In addition, to evaluate the overall anisotropy of the embedding space, we compute the average cosine similarity of 10,000 randomly sampled sentence pairs.
\subsection{The global change in the hidden representation space} We observe two complementary effects of SFT on the model’s embedding space. First, the number of identifiable clusters decreases \ref{fig:number_of_cluster}: representations that were once scattered into many small groups collapse into a smaller set of semantically coherent modes, indicating that the model has learned to emphasize only those coarse distinctions that are most relevant for the downstream task. Second, the ID of each remaining cluster increases \ref{fig:ID_all}: within each merged mode, embeddings spread out along additional directions, reflecting the model’s acquisition of subtler, task-specific features. Together, these trends suggest a trade-off in which fine-tuning simplifies the global structure of the representation (fewer clusters) while enriching its local expressiveness (higher ID), thereby balancing coarse category separation with finer-grained feature encoding.

\ref{fig:ARI_all} shows that ARI fluctuates markedly across layers in every model, highlighting its sensitivity to representational changes. Fine-tuned variants generally exhibit lower ARI than their pretrained counterparts, indicating that clustering consistency does not directly predict generative performance. Moreover, because ARI here is computed over the full set of training-set embeddings, its overall trend may obscure differences between in-distribution and out-of-distribution samples. To disentangle these effects, we next perform an analysis of embedding–dataset correspondence.
\subsection{The local change in the hidden representation space} As we show in the Figure \ref{fig:number_of_cluster}, the number of clusters tends to be smallest in the final layer. We interpret this as the model forming semantically meaningful groupings in the embedding space at the final layer. Therefore, we examine the breakdown of hidden representations in the final layer of OLMo-7B.
To evaluate the properties of the hidden representations for both in-distribution and out-of-distribution training datasets, we compute kNN consistency as defined by the following equation.

\begingroup
    \setlength{\abovedisplayskip}{6pt}
    \setlength{\belowdisplayskip}{6pt}
    \footnotesize
    \begin{equation*}
    \mathrm{kNN\, consistency}_c := 
    \frac{1}{N_c} \sum_{i: y_i = c}
    \left[
    \frac{1}{k} \sum_{j \in \mathcal{N}_k(i)} 
    \mathbb{I}(y_j = y_i)
    \right]
    \end{equation*}
\endgroup

\begin{figure*}[t]
    \centering
    \includegraphics[width=\linewidth]{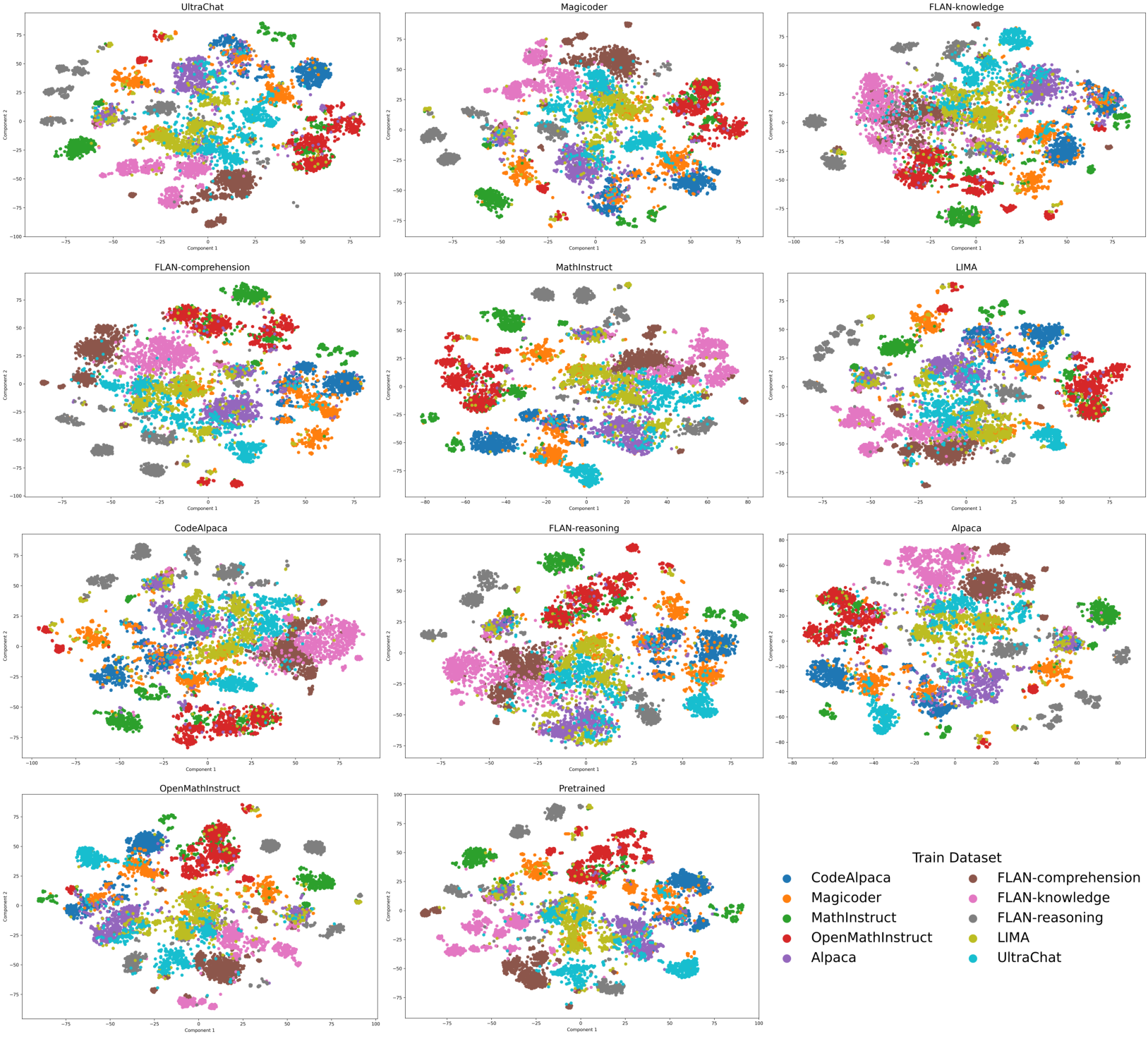}
    \caption{\textbf{t-SNE visualization of OLMo-7B at last layer.} As an overall trend, the hidden representations of the trained (in-distribution) dataset become more tightly clustered, while the representations of the untrained (out-of-distribution) datasets show reduced discriminability and their distributions become more mixed with those of other datasets.}
    \label{fig:T-SNE_OLMo}
\end{figure*}

Let $y_i$ be the label of the $i$-th hidden representation in the training dataset, where $i \in {1, 2, \ldots, N}$. For each embedding $i$, let $\mathcal{N}_k(i)$ denote the set of its $k$-nearest neighbors in the hidden representation space. To understand how well the local neighborhood of each data point matches its label, we calculate the kNN consistency for each label $c$. Specifically, for each data point whose label is $c$, we compute the proportion of its $k$ nearest neighbors that have the same label. We then average this value over all points with label $c$:
A label consistency of 1.0 means every point’s neighbors are all the same class (perfect local purity), whereas lower values signify that points are often neighbored by different classes. In our experiments, we set $k=300$.
\begin{figure}[H]
    \centering
    \includegraphics[width=\linewidth]{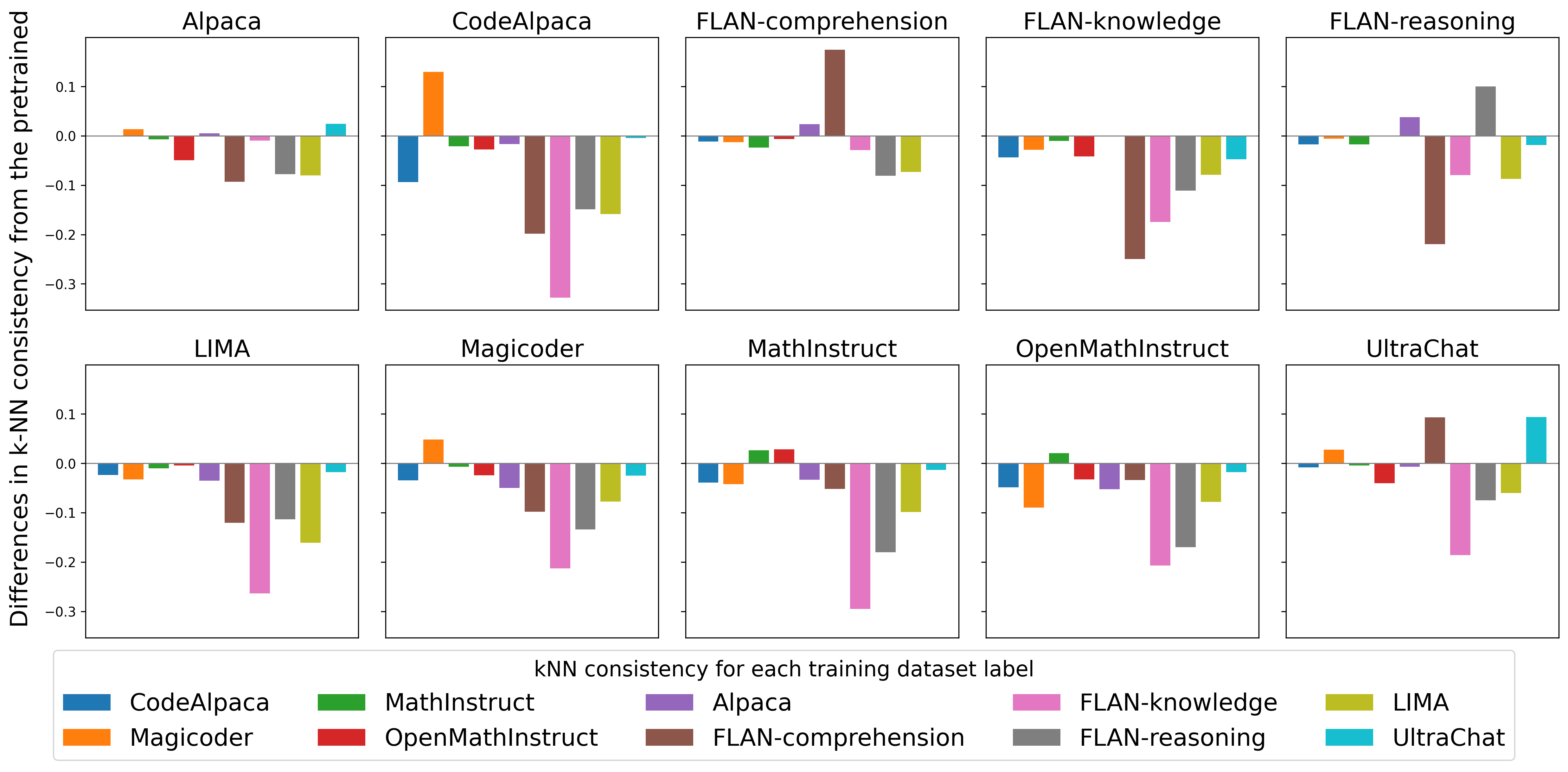}
    \caption{\textbf{Differences in kNN consistency from the pretrained model for OLMo-7B.} This shows how the kNN consistency in the final layer changes for each dataset label when OLMo-7B is fine-tuned on 1k examples from each training dataset. For example, when OLMo-7B is fine-tuned on 1k examples from CodeAlpaca, it becomes better at embedding sentences from Magicoder more closely. On the other hand, for other datasets, the pretrained model demonstrates better separability.}
    \label{fig:OLMo-kNN}
\end{figure}
Figure \ref{fig:OLMo-kNN} shows how kNN consistency of the embedding representations changes due to fine-tuning. Intuitively, the embeddings of datasets that belong to the same category as the training dataset—i.e., in-distribution—tend to become more tightly clustered, while embeddings of out-of-distribution datasets become harder to distinguish. In practice, when fine-tuned on FLAN-comprehension, FLAN-reasoning, or Magicoder, we observed a decrease in kNN consistency for datasets other than the one used for training. Similarly, when fine-tuned on MathInstruct, kNN consistency decreased for all datasets except MathInstruct and OpenMathInstruct. This phenomenon is illustrated in Figure \ref{fig:T-SNE_OLMo} by projecting the embedding space into two dimensions using t-SNE. The pretrained model produces many small clusters, but it can still distinguish the labels of the training datasets. In contrast, the models fine-tuned on each training dataset show embeddings that are more tightly clustered together, making it more difficult to distinguish between the dataset labels. The mechanism of unlearning is likely caused by the model's embedding representations becoming less distinguishable for out-of-distribution datasets. Therefore, it will be beneficial to train on low-perplexity datasets that do not deviate too far from the base model's original distribution.

%========================================%
% \section{Result}
% \label{sec: Raw results}

% \begin{table}[ht]
%   \centering
%   \csvautotabular{tables/result_simple.csv}
% \end{table}

% \begin{table}[ht]
%   \centering
%   \csvautotabular[command=\texttt]{tables/result_simple1.csv}
% \end{table}

% \begingroup
% \tiny
% \catcode`\_=12
% \csvautotabular{tables/result_simple1.csv}
% \endgroup

% \onecolumn
% {\scriptsize
% % 1. 横幅をページに収まるよう自動調整
% \resizebox{\textwidth}{!}{%
% \begin{longtable}{|l|l|l|l|l|l|l|l|l|l|l|l|l|l|l|l|l|l|l|}
% \hline
% Model & Dataset & lora & epoch & data\_size & lr\_strategy & MATH & HumanEval & MT-Bench & AlpacaEvalv2 & MBPP & GSM8K & NQ & BoolQ & Hellaswag & MMLU & TrutufulQA & JMMLU & CMMLU\\
% \hline
% \endfirsthead
% \hline
% Model & Dataset & lora & epoch & data\_size & lr\_strategy & MATH & HumanEval & MT-Bench & AlpacaEvalv2 & MBPP & GSM8K & NQ & BoolQ & Hellaswag & MMLU & TrutufulQA & JMMLU & CMMLU\\
% \hline
% \endhead
% \csvreader[head to column names]{tables/result_simple.csv}{}%
% {%
% \csvcoli & \csvcolii & \csvcoliii & \csvcoliv & \csvcolv & \csvcolvi & \csvcolvii & \csvcolviii & \csvcolix & \csvcolx & \csvcolxi & \csvcolxii & \csvcolxiii & \csvcolxiv & \csvcolxv & \csvcolxvi & \csvcolxvii & \csvcolxviii & \csvcolxix\\
% \hline
% }
% \end{longtable}
% }
% }
% \twocolumn

% \pgfplotstableread[col sep=comma]{tables/result_simple.csv}\datatable
% \pgfplotstabletypeset{\datatable}

\end{document}